\documentclass[runningheads]{llncs}

\usepackage{eccv}

\usepackage{eccvabbrv}

\usepackage{graphicx}
\usepackage{booktabs}

\usepackage[accsupp]{axessibility}  % Improves PDF readability for those with disabilities.
\usepackage{hyperref}

\usepackage{orcidlink}

\usepackage{booktabs}
\usepackage{multirow}

\usepackage{array}
\usepackage{pifont}
\usepackage{verbatim}
\usepackage{array}
\usepackage[table]{xcolor} % for \rowcolor
\usepackage{colortbl}
\usepackage{xcolor}
\usepackage{subcaption}
\usepackage{multibib}
\usepackage{graphicx}

\begin{document}

\newcommand{\approach}{Co-VGGT}
\newcommand{\fz}[1]{\textcolor{red}{#1}}

\title{What VGGT Knows About Overlap: Probing Geometric Foundation Models for Co-Visibility}


\titlerunning{What VGGT Knows About Overlap}

\author{Filippo Ziliotto\inst{1,2}\orcidlink{0009-0003-0466-2423} \and
Luciano Serafini\inst{2}\orcidlink{0000-0003-4812-1031} \and 
Lamberto Ballan\inst{1}\orcidlink{0000-0003-0819-851X} \and
Tommaso Campari\inst{2}\orcidlink{0000-0002-0435-4397}}

%\authorrunning{F.~Author et al.}
\authorrunning{F. Ziliotto et al.}

\institute{University of Padova \and
Fondazione Bruno Kessler (FBK)
}

\maketitle

\newcommand{\tc}[1]{\textcolor{blue}{#1}}
\newcommand{\cmark}{\ding{51}}%
\newcommand{\xmark}{\ding{55}}%

\begin{abstract}
A fundamental challenge in 3D reconstruction and robotic localization is co-visibility: determining which image pairs share overlapping visible surfaces, particularly in scenarios with minimal overlap.
We demonstrate that VGGT implicitly encodes co-visibility as an emergent behavior: without any supervision for this task, its internal representations exhibit a clear hierarchical structure mirroring that of large language models, i.e. early layers build a 3D-aware scene representation, while late layers act as dedicated co-visibility reasoners.
In particular, we identify layer L17 as a negative anchor that consistently routes non-co-visible pairs for this backbone, regardless of the evaluation setting, providing task-grounded evidence of layer specialization in a geometry-grounded foundation model.
Building on this, we introduce \approach, which freezes VGGT and trains only a lightweight layer-wise mixture-of-experts head (${\sim}7.5$M parameters) to classify co-visibility from RGB alone, treating each layer as a specialized expert whose geometric abstraction is adaptively weighted per input pair.
On the Co-VisiON benchmark, \approach\ surpasses the human annotation baseline and improves over prior work by more than 25\% pairwise and 10\% multiview.
Pairwise predictions are well-calibrated (ECE\,$=$\,0.030), enabling direct use as edge weights in visibility graphs for downstream SfM and SLAM pipelines without post-hoc correction. Code and data are available\footnotemark[1]. \footnotetext[1]{ \href{https://github.com/filippoziliotto/covisibility-probing/}{https://github.com/covisibility-probing}}
\end{abstract}

%%%%%%%%%%%%%%%%%%%%%%%%%%%%%%%%%%%%%%%%%%%%%%%%%%%%%%%%%%%%%%%%%%%%%%%%%%%%%%%%

\section{Introduction}
\label{sec:introduction}

\begin{figure}[th]
  \centering
  \includegraphics[width=\columnwidth]{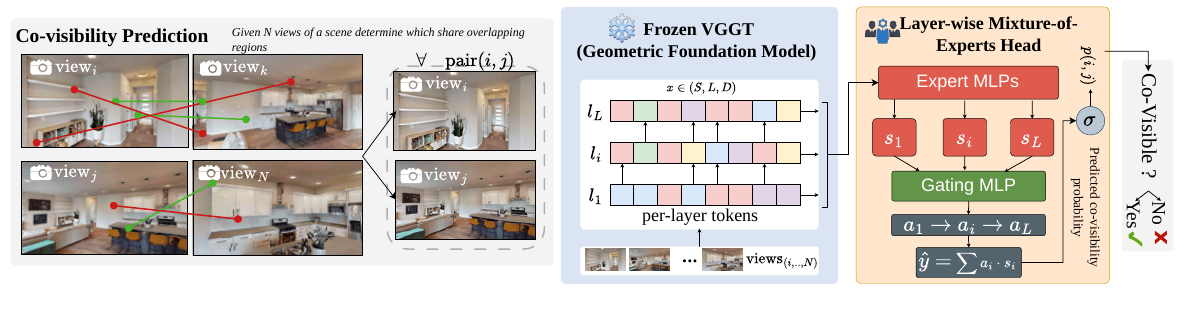}
    \caption{\textbf{Co-visibility Task.} We study co-visibility prediction: given multiple RGB views of a scene, the goal is to determine which image pairs share overlapping visible regions. Our approach probes geometric consistency in a frozen VGGT foundation model, extracting layer-wise view embeddings and combining them through a lightweight mixture-of-experts head. Without modifying the backbone, the model aggregates information across transformer layers to produce a co-visibility probability for each pair, enabling efficient construction of scene-level visibility graphs.}
  \label{fig:teaser}
\end{figure}

Robotic perception and 3D reconstruction systems typically operate on sparse, imperfect image sets rather than isolated, perfect views. 
A fundamental challenge in these pipelines—whether for mapping, localization, or scene reconstruction—is determining co-visibility, the subset of surfaces jointly observed by multiple cameras. 
High co-visibility yields abundant geometric constraints and stable optimization. 
Conversely, when spatial overlap is limited or absent, matching becomes ambiguous and pose estimation drifts. 
In these scenarios, reconstruction pipelines often fail silently, generating plausible but geometrically inconsistent structures. 
Operating in this sparse-view regime is not an edge case but rather a standard condition for embodied agents navigating complex, occluded environments.

Despite advances in multiview transformers and learned reconstruction models, performance drops significantly as spatial overlap decreases.
Under these conditions, fusion modules misalign features, learned descriptors drift, and global reasoning degrades into spurious correlations.
Recent benchmarks, such as Co-VisiON~\cite{chen2025covision}, explicitly isolate this co-visibility reasoning, exposing a critical performance gap between current models and human baselines in sparse, highly imbalanced scenarios.
Bridging this gap is essential for robotics: robust co-visibility estimation dictates which view pairs to match, which constraints to trust, and when to flag reconstruction failures.

Concurrently, geometry-grounded foundation models present a promising alternative. 
The Visual Geometry Grounded Transformer (VGGT)~\cite{wang2025vggt} demonstrates strong emergent geometric reasoning; without explicit 3D supervision, it infers scene structure and reconstructs geometry even across widely separated viewpoints. 
Similar to emergent phenomena in large language models, VGGT's internal representations encode latent structures that transcend its immediate training objective. 
However, the exact nature of these spatial reasoning signals and the methodology to extract them for embodied decision-making remain poorly understood.

In this work, we show that frozen VGGT features contain a strong, directly usable signal for co-visibility estimation (see Fig.~\ref{fig:teaser}).
Furthermore, we find that specific late VGGT layers act as consistent anchors for co-visibility decisions, while earlier layers encode broader scene-aware geometric representations.
Leveraging this property, we propose Co-VGGT, an MoE head in which each layer acts as an ``expert'' whose geometric abstraction is adaptively weighted per image pair. Compared to single-layer probing, this routing improves accuracy and calibration while exposing which layer cues drive each decision.
Relying solely on RGB inputs, Co-VGGT achieves near human-level accuracy on the Co-VisiON benchmark.
It surpasses the current state-of-the-art by over 25\% in pairwise co-visibility prediction and nearly 10\% in multiview inference.
We further observe that co-visibility reasoning is dominated by late layers, whereas earlier layers contribute more general geometric context.

Our findings indicate that geometry-grounded foundation models encode spatial priors that are far richer than those obtained through conventional contrastive or matching-based pretraining.
Furthermore, these signals can be efficiently distilled into modular predictors.
Such predictors could be integrated into reconstruction pipelines: a dedicated co-visibility head can refine overlap prediction and serve as an auditing mechanism, identifying inconsistent constraints and detecting early-stage failures to support reliable embodied autonomy.

In summary, our contributions are as follows: (i) we introduce Co-VGGT, achieving near human-level co-visibility from RGB and improving the Co-VisiON SOTA by >25\% (pairwise) and  $\sim$10\% (multiview); (ii) we provide evidence that VGGT implicitly encodes co-visibility structure across its layers, exposing emergent spatial priors that are absent in standard supervised baselines; and (iii) we show that Co-VGGT is robust to minimal overlap scenarios and its well-calibrated outputs (ECE\,$=$\,0.030 on Gibson val) enable direct use as edge weights in visibility graphs, providing a practical auditing signal for SfM and SLAM pipelines without post-hoc correction.

\section{Related Works}
\label{sec:related_work}

Co-visibility prediction sits at the intersection of 3D reconstruction, feature matching, and geometric reasoning. We review the most relevant lines of work, highlighting how each studies co-visibility as an explicit, predictable signal. 
In doing so, we ground our approach within the broader landscape and clarify what makes it fundamentally distinct from all prior work.

\noindent\textbf{Co-visibility in Reconstruction and Matching Pipelines.}
Co-visibility is handled implicitly in classical SfM and SLAM through feature matching and geometric verification~\cite{schonberger2016structure, campos2021orbslam3}, degrading under weak texture or large viewpoint changes. Learned SLAM methods~\cite{teed2021droidslam, zhang2021badslam} and implicit neural mapping~\cite{sucar2021imap, zhu2022niceslam} improve robustness but still require sufficient overlap, while systems with explicit graph structures — Kimera~\cite{rosinol2020kimera}, Hydra~\cite{hughes2022hydra}, and situational graphs~\cite{bavle2022sgraph} — treat co-visibility as a derived post-hoc quantity. Learned matchers — SuperGlue~\cite{sarlin2020superglue}, LightGlue~\cite{lindenberger2023lightglue}, Efficient LoFTR~\cite{wang2024efficientloftr}, and OmniGlue~\cite{jiang2024omniglue} — estimate matchability conditioned on the existence of overlap and cannot flag non-overlapping pairs, while retrieval methods like NetVLAD~\cite{arandjelovic2016netvlad} conflate semantic similarity with geometric overlap. We instead predict co-visibility directly from RGB as a dedicated, calibrated first-class signal before any reconstruction is attempted.

\noindent\textbf{Geometric Foundation Models and Sparse-View Reasoning.}
DUSt3R~\cite{wang2024dust3r}, MUSt3R~\cite{cabon2025must3r}, MVSNeRF~\cite{chen2021mvsnerf}, and MV-DUSt3R+~\cite{xu2023mvfusion} have advanced multiview geometry estimation; yet, all degrade in sparse-view regimes where co-visible surface support is limited. VGGT~\cite{wang2025vggt} achieves emergent 3D reasoning without explicit geometric supervision; we show its internal representations encode a hierarchically-organized co-visibility signal that can be distilled without modifying the backbone. The Co-VisiON benchmark~\cite{chen2025covision} formalizes this as graph inference over sparse indoor views, exposing a critical performance gap that our approach substantially closes.

\noindent\textbf{VLMs, Geometric Distillation, and Transformer Interpretability.}
VLMs such as GPT-4o~\cite{openai2023gpt4}, Gemini~\cite{google2023gemini}, and SpatialRGPT~\cite{cheng2024spatialrgpt} show emergent spatial reasoning but remain unreliable for precise 3D consistency under significant viewpoint changes, as confirmed by our experiments. Recent distillation approaches inject geometric priors into frozen language backbones~\cite{lee2025geometricdistillation, an2025evggt}, but target general scene understanding rather than co-visibility. Meanwhile, probing work localizes capabilities to specific transformer layers~\cite{tenney2019bert, geva2021s, stary2025understanding}, with reasoning concentrated in the late layers of LLMs~\cite{meng2022locating} and global semantics in the late layers of vision transformers~\cite{xue2022protopformer}. We provide the first task-grounded evidence of an analogous hierarchical structure in a geometry-grounded model, identifying the late VGGT layers (particularly L17) as decisive co-visibility reasoners.

\section{Method}
\label{sec:method}

\begin{figure}[t]
  \centering
  \includegraphics[width=\columnwidth]{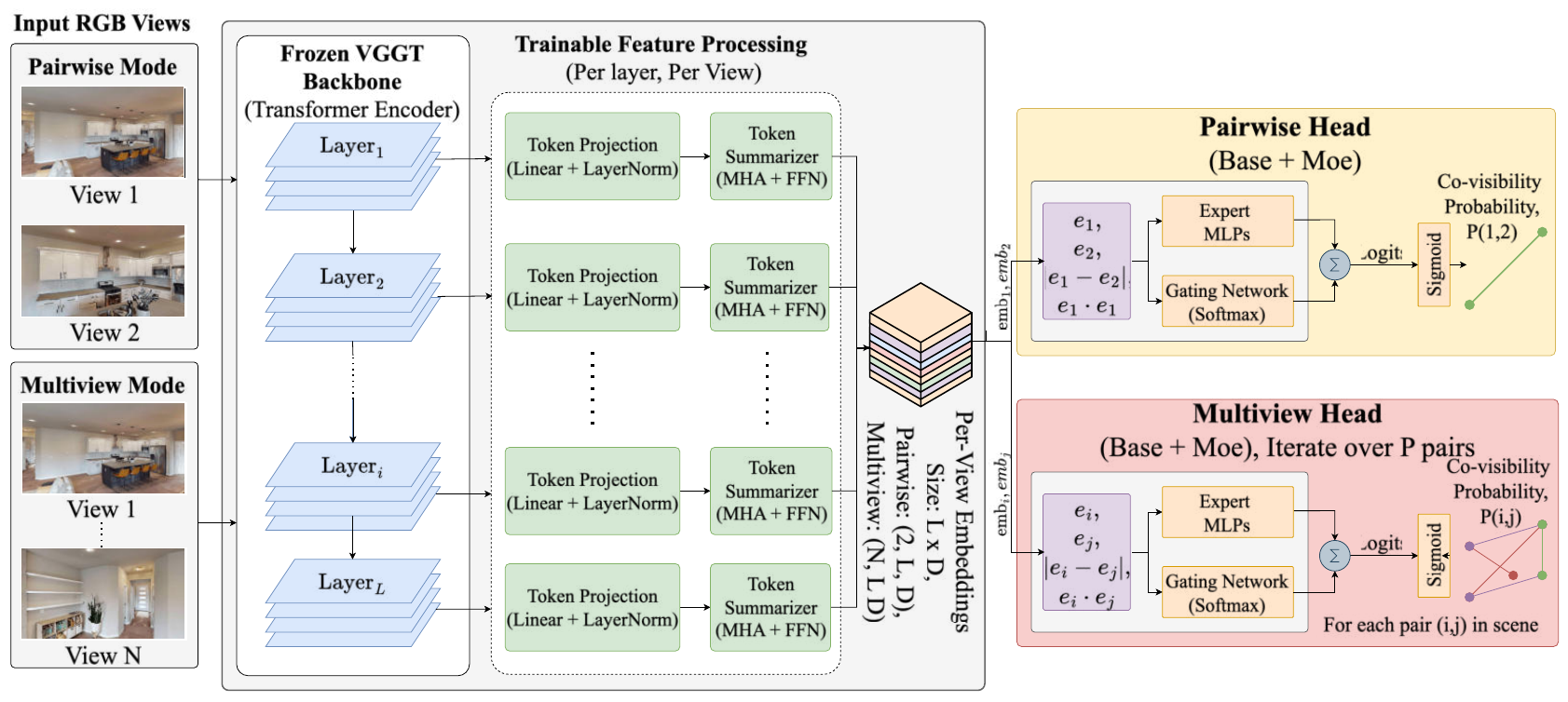}
    \caption{\textbf{Overview of the Co-VGGT method}. Input RGB views are processed by the frozen VGGT backbone to extract layer-wise features. These features are projected, summarized, and formed into per-view embeddings. In both pairwise and multiview modes, these embeddings are used to construct pair features, which are then fed into a trainable Mixture-of-Experts (MoE) head to predict co-visibility probabilities, enabling the construction of scene-level visibility graphs. Tensor shapes are annotated for key intermediate representations.}
  \label{fig:method}
\end{figure}

We address co-visibility estimation simply using RGB inputs by training a lightweight binary classifier on top of a frozen geometric foundation model. Given a set of views from the same scene, the goal is to predict whether each view pair shares any jointly visible surface region to reconstruct the scene-graph. Our model operates in two regimes:
(i) pairwise, where each training example is a single pair of images; and
(ii) multiview, where each example is an entire scene containing many views and labeled pairs. 
An overview of the method is shown in Fig.~\ref{fig:method}.

Importantly, pairwise inference is a special case of multiview with $N{=}2$ and a single pair.
This unified formulation allows training in multiview mode and evaluation in both pairwise and multiview settings without architectural changes.

\noindent\textbf{Problem Setup.\ }
Let $\mathcal{I}=\{I_s\}_{s=1}^{S}$ be a set of RGB views of the same scene, with $I_s\in\mathbb{R}^{3\times H\times W}$.
For any pair $(i,j)$, the target is a binary co-visibility label $y_{ij}\in\{0,1\}$.
In the pairwise setting we predict a single labeled pair per sample ($S{=}2$).
In the multiview setting we process $S{>}2$ views jointly and predict a set of labeled pairs $\mathcal{P}=\{(i_p,j_p,y_p)\}_{p=1}^{P}$, which defines a co-visibility graph over the views.

\noindent\textbf{Backbone and Summarization.\ }
We use a pretrained VGGT backbone $\Phi$ as a frozen feature extractor.
Given a batch $\mathbf{X}\in\mathbb{R}^{B\times S\times 3\times H\times W}$, we extract patch-token features from selected layers $\ell\in\{1,\dots,L\}$:
$\mathbf{T}^{(\ell)}=\Phi^{(\ell)}(\mathbf{X})\in\mathbb{R}^{B\times S\times P_{\text{tok}}\times C_{\text{raw}}}$.
All parameters of $\Phi$ remain frozen.
To obtain compact per-view embeddings, we (i) project token channels and (ii) summarize tokens with learned queries.
We first apply a shared linear projection with LayerNorm:
$\widehat{\mathbf{T}}^{(\ell)}=\mathrm{LN}\!\left(\mathbf{T}^{(\ell)}W\right)$, where $W\in\mathbb{R}^{C_{\text{raw}}\times C_{\text{proj}}}$.
Then, for each view we summarize $P_{\text{tok}}$ tokens into $T$ summary tokens via cross-attention with learned queries $\mathbf{Q}\in\mathbb{R}^{T\times C_{\text{proj}}}$:
\begin{equation}
\mathbf{S}^{(\ell)}_{s}=\mathrm{Attn}\!\left(\mathbf{Q},\widehat{\mathbf{T}}^{(\ell)}_{s},\widehat{\mathbf{T}}^{(\ell)}_{s}\right)\in\mathbb{R}^{T\times C_{\text{proj}}},
\quad
\mathbf{e}^{(\ell)}_{s}=\mathrm{vec}\!\left(\mathbf{S}^{(\ell)}_{s}\right)\in\mathbb{R}^{D},
\label{eq:summarizer}
\end{equation}
with $D=T\,C_{\text{proj}}$. This yields per-view, per-layer embeddings $\{\mathbf{e}^{(\ell)}_{s}\}_{\ell=1}^{L}$.

\begin{figure}[t]
  \centering
  \includegraphics[width=\columnwidth]{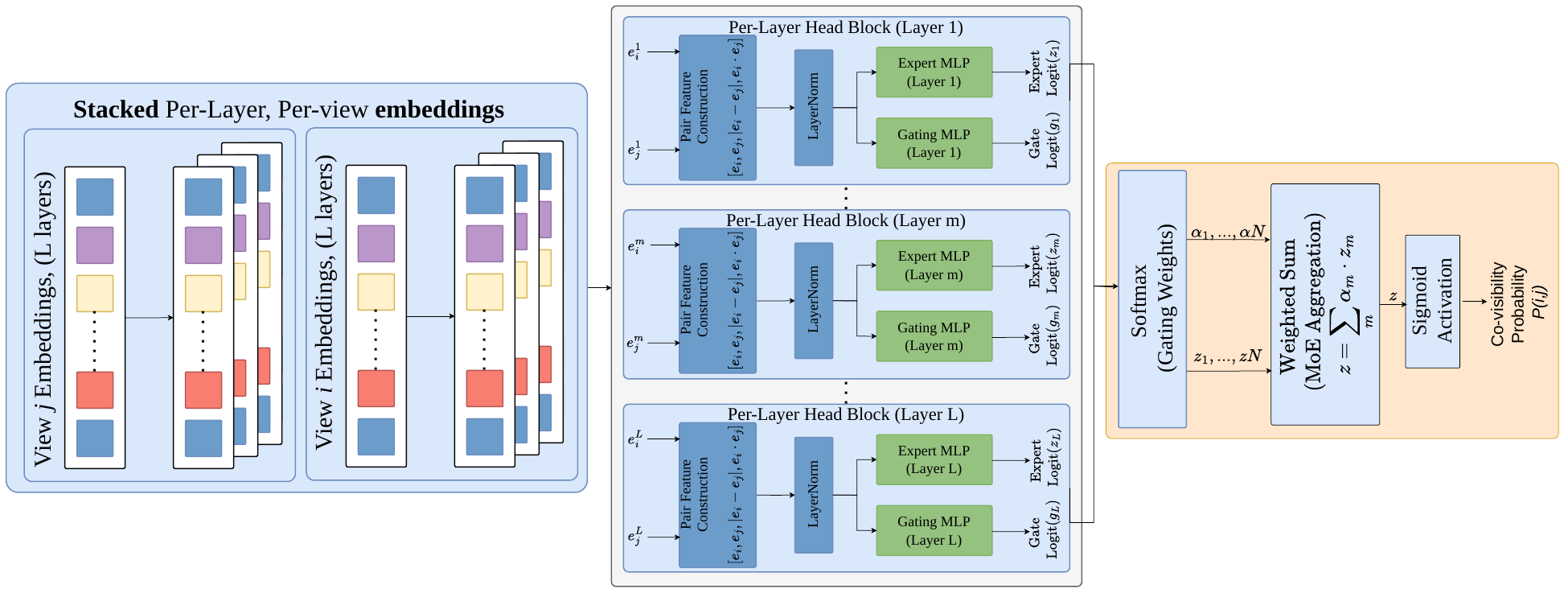}
\caption{\textbf{Co-visibility Estimation Head.} Given a set of sparse RGB views (left), our method leverages the frozen Visual Geometry Grounded Transformer (VGGT) to extract layer-wise view embeddings. A lightweight, trainable Mixture-of-Experts (MoE) head (center) adaptively aggregates these multi-scale features to predict pairwise co-visibility probabilities. The resulting predictions form a dense scene-level visibility graph (right), identifying overlapping regions between disjoint viewpoints to guide robust 3D reconstruction and robotic perception.}
  \label{fig:method_head}
\end{figure}

\noindent\textbf{Pair Representation \& MoE Head.\ }
Following \cite{mou2016natural, grover2016node2vec, conneau2017supervised}, to enhance the embedding representation for each labeled pair $(i,j)\in\mathcal{P}$ and layer $\ell$, we build a symmetric pair feature:
\begin{equation}
\mathbf{f}^{(\ell)}_{ij}=
\left[
\mathbf{e}^{(\ell)}_{i},\;
\mathbf{e}^{(\ell)}_{j},\;
\left|\mathbf{e}^{(\ell)}_{i}-\mathbf{e}^{(\ell)}_{j}\right|,\;
\mathbf{e}^{(\ell)}_{i}\odot \mathbf{e}^{(\ell)}_{j}
\right]\in\mathbb{R}^{4D}.
\label{eq:pair_feature}
\end{equation}
Based on the assumption that VGGT mirrors the hierarchical reasoning structure of LLMs, in our method, each layer acts as an ``expert'' that predicts a logit $z^{(\ell)}_{ij}$ from $\mathbf{f}^{(\ell)}_{ij}$, while a gating network assigns mixture weights across layers:
\begin{equation}
z^{(\ell)}_{ij}=\mathrm{MLP}_{\text{exp}}\!\left(\mathrm{LN}(\mathbf{f}^{(\ell)}_{ij})\right),
\quad
\alpha^{(\ell)}_{ij}=\mathrm{softmax}_{\ell}\!\left(\mathrm{MLP}_{\text{gate}}(\mathrm{LN}(\mathbf{f}^{(\ell)}_{ij}))\right).
\label{eq:moe_expert_gate}
\end{equation}
The final logit and probability are
\begin{equation}
z_{ij}=\sum_{\ell=1}^{L}\alpha^{(\ell)}_{ij}\,z^{(\ell)}_{ij},
\qquad
p_{ij}=\sigma(z_{ij}),
\label{eq:moe_output}
\end{equation}
as illustrated in Fig.~\ref{fig:method_head}.

\noindent\textbf{Training Objective.\ }
We optimize the projector, summarizer, expert, and gate parameters with binary cross-entropy on logits:
\begin{equation}
\mathcal{L}=\frac{1}{|\mathcal{P}|} \sum_{(i,j,y_{ij})\in\mathcal{P}} \ell_{\mathrm{BCELogits}}(z_{ij}, y_{ij}).
\end{equation}
In pairwise mode $|\mathcal{P}|{=}1$, while in multiview mode, we average over all labeled pairs in the scene. 
Although co-visibility resembles image similarity, naive embedding matching is brittle under viewpoint changes, occlusion, and texture ambiguity. 
This structure empirically reduces false positives in low-overlap regimes while improving calibration.

\subsection{Zero-Shot Evaluation}
\label{sec:zero_shot}

We introduce a zero-shot baseline that predicts co-visibility without any trainable parameters. 
The classification head and mixture-of-experts are removed, and scores are computed directly from frozen VGGT features using cosine similarity.
Given a pair of RGB images, we extract layer activations $\{\mathbf{T}^{(\ell)}\}_{\ell=1}^{L}$ from the frozen backbone and apply a pooling operator $\mathcal{P}(\cdot)$ to obtain per-view embeddings $\mathbf{e}^{(\ell)}\in\mathbb{R}^{D}$. 
Embeddings are then computed for each of the considered layers (see Tab.~\ref{tab:zeroshot_ablation} for details).

For a pair $(i,j)$, we compute the cosine similarity per layer and average across layers:
\begin{equation}
    c_{ij} = \frac{1}{L}\sum_{\ell=1}^{L} 
\frac{\langle \mathbf{e}^{(\ell)}_i, \mathbf{e}^{(\ell)}_j \rangle}
{\|\mathbf{e}^{(\ell)}_i\|_2 \|\mathbf{e}^{(\ell)}_j\|_2} \in [-1,1].
\end{equation}

Finally, we rescale to $[0,1]$ to obtain a pseudo-probability 
$p^{\text{ZS}}_{ij} = (c_{ij}+1)/2$, 
which is evaluated using the same metrics as the trained models.

\section{Experiments}
\label{sec:experiments}

We evaluate our method on the Co-VisiON benchmark across both HM3D and Gibson environments. 
We report results for the \emph{pairwise} and \emph{multiview} settings and include the zero-shot method described in Sec.~\ref{sec:method}.
All experiments use AdamW with a learning rate $10^{-4}$, a batch size of 32, and a weight decay $10^{-4}$, trained for 50 epochs using the results corresponding to the best AUC value.
In the multiview setting, peak performance is reached around 30 epochs, whereas in the pairwise setting, it converges within 10 epochs.

\subsection{Dataset \& Metrics}
Co-VisiON~\cite{chen2025covision} evaluates co-visibility reasoning in sparse indoor view sets. 
Each sample consists of a small collection of RGB views from the same scene, and the task is to predict a binary co-visibility graph, where an edge $(i,j)$ is positive iff the two views share non-zero co-visible surface area. 
%Labels are generated in simulation via pixel-level visibility rendering on textured meshes, ensuring accurate supervision without reconstructed geometry.
The benchmark spans Gibson~\cite{xia2018gibson} (80/20 split) and HM3D~\cite{ramakrishnan2021hm3d} (90/10 split), comprising 85/755 scenes and 33,849/210,008 labeled pairs, respectively.
Moreover, all validation scenes are disjoint from training scenes, so performance reflects generalization to unseen environments, akin to a zero-shot setting.
%In both cases, the training and test scenes are disjoint and depict different indoor environments, allowing us to explicitly assess the model’s generalization capability.
For detailed specifics, refer to \cite{chen2025covision}.

The dataset also reports a human baseline for the Gibson multiview setting. However, this should not be interpreted as a definitive measure of human-level performance, but rather as an indicative reference point. Given the presence of artifacts in Gibson, the true performance achievable by humans under cleaner conditions is likely higher.

We use the same protocol for \emph{pairwise} and \emph{multiview} settings since both produce pairwise co-visibility scores that define an adjacency matrix. 
For a scenario with $N$ views, the model outputs scores $d_{ij}\in[0,1]$, forming $\mathbf{D}\in[0,1]^{N\times N}$, which are compared to the ground-truth adjacency $\mathbf{A}\in\{0,1\}^{N\times N}$. 
The only difference is how scores are computed: independently per pair in the pairwise setting, or jointly from multiple views in the multiview setting.

Given a threshold $\tau$, we binarize $\mathbf{D}$ as $\hat{\mathbf{A}}(\tau)=\mathbb{I}[\mathbf{D}\ge\tau]$ (symmetric, zero diagonal) and compute the \emph{Graph IoU}:
$\mathrm{IoU}(\tau)=\frac{|\mathcal{E}\cap \hat{\mathcal{E}}(\tau)|}{|\mathcal{E}\cup \hat{\mathcal{E}}(\tau)|}$.
We report the best-threshold IoU (denoted as $\text{IoU}^*$ in the tables), $\mathrm{IoU}^{*}=\max_{\tau\in[0,1]}\mathrm{IoU}(\tau)$, and the area under the IoU-threshold curve over $\tau\in[0,1]$, defined as
$\mathrm{AUC}=\int_{0}^{1}\mathrm{IoU}(\tau)\, d\tau$.
Metrics are computed per scenario and averaged over the split.
Additional metrics results (e.g., F1 score, Accuracy) are provided in Sec.~A.6 of the Supplementary material.

\subsection{Results}
\label{sec:results}

Tabs.~\ref{tab:covis_pairwise}--\ref{tab:covis_multiview} report co-visibility prediction performance on Co-VisiON in the pairwise and multiview settings, measured with Graph IoU* and AUC. 
Our \approach\ consistently achieves the best results across datasets and evaluation protocols. 
In the pairwise setting (Tab.~\ref{tab:covis_pairwise}), \approach\ attains 0.85/0.78 IoU*/AUC on Gibson and 0.84/0.78 on HM3D, substantially outperforming strong learned baselines such as Covis~\cite{chen2025covision} (0.56/0.54 on Gibson; 0.53/0.51 on HM3D) and reconstruction based DUSt3R~\cite{wang2024dust3r} (0.54/0.54 on Gibson; 0.40/0.40 on HM3D). 
Prompting large VLMs yields competitive results relative to earlier learned models (e.g., GPT-4o~\cite{openai2023gpt4} reaches 0.58/0.58 on Gibson and 0.54/0.54 on HM3D), but remains far behind \approach, indicating that explicit geometric specialization is critical for reliable overlap reasoning under sparse viewpoints. 
%In the pairwise setting (Tab.~\ref{tab:covis_pairwise}), Co-VGGT attains 0.85/0.78 IoU/AUC on Gibson and 0.84/0.78 on HM3D, well above the strongest learned baseline (Covis~\cite{chen2025covision}), reconstruction-based DUSt3R [35], and prompted VLMs (GPT-4o~\cite{openai2023gpt4} is the best at 0.58/0.58 Gibson) — indicating that explicit geometric specialization is critical for overlap reasoning under sparse viewpoints.

In the multiview setting (Tab.~\ref{tab:covis_multiview}), \approach\ again leads with 0.74/0.72 on Gibson and 0.76/0.74 on HM3D, surpassing Covis/Covis-freeze and multiview reconstruction (MV-DUSt3R+~\cite{tang2024mvdust3r}). 
Interestingly, performance in multiview is lower than in pairwise, which is counterintuitive for reconstruction-style models but can be explained by our current multiview embedding extraction, which aggregates features across many views and yields noisier per-view vectors for the downstream pair classifier. 

%Moreover, the model architecture is fundamentally pairwise: it scores each view pair independently, even in multiview scenarios, compressing all multiview context into fixed-dimensional pair embeddings. This design lacks explicit graph-level reasoning mechanisms, potentially limiting its ability to enforce consistency across the full co-visibility graph, leaving space for improvements.
On Gibson, \approach\ also exceeds the reported Human Annotation baseline (0.72/0.72)~\cite{chen2025covision}, suggesting that the learned head on top of frozen VGGT features captures geometric cues that are difficult to assess reliably from sparse RGB images presented to humans. 

We note that for sparse view sets (small $N$), running the pairwise model exhaustively over all $\binom{N}{2}$ pairs and assembling the graph post-hoc yields superior calibration and accuracy (0.85 vs.\ 0.74 IoU*) at negligible additional cost, and is therefore the recommended inference strategy in practice. 
The multiview mode offers a scalable alternative for larger view sets where the quadratic cost becomes prohibitive.

Tab.~\ref{tab:difficulty_levels} reports validation performance under two complementary difficulty protocols that stress different sparsity regimes of the co-visibility graph. 
Edge-level difficulty bins individual pairs by their overlap ratio: \textit{easy} if overlap $\geq 50\%$, \textit{medium} if $10\%\leq$ overlap $<50\%$, and \textit{hard} if overlap $<10\%$. 
While VLM-based baselines are near-saturated in the easy/medium regimes, they degrade sharply when overlap becomes minimal; in particular, on the hard split \approach\ achieves substantially higher Graph-IoU than the strongest VLM baseline (e.g., $0.84$ vs.\ $0.34$ for GPT-4o~\cite{openai2023gpt4}), indicating markedly better robustness to low-overlap pairs. 
We further report graph-level difficulty by binning entire scenes according to scene sparsity, defined as the average pairwise overlap within the scenario: \textit{easy} if $\geq 10\%$, \textit{medium} if $4\%\leq\cdot<10\%$, and \textit{hard} if $<4\%$. 
Even in globally sparse scenes, \approach\ maintains strong performance (hard: $0.84$ Graph-IoU), outperforming both Covis-based baselines and GPT-4o (hard: $0.57$), showing that the method remains reliable when the overall graph connectivity is low, rather than just a few pairs being difficult. 
Overall, these results suggest that \approach\ encodes more stable geometric cues than VLM prompting or learned baselines, and that its advantage concentrates exactly where co-visibility is most failure-prone: extremely low pairwise overlap and scene-wide sparse connectivity.
AUC result tables are reported in the Supplementary material.

%Finally, the zero-shot variant of our pipeline—based solely on the cosine similarity of frozen VGGT embeddings—already yields non-trivial performance (e.g., 0.55 IoU* on the Gibson pairwise), but falls well short of the trained model in both pairwise and multiview regimes, confirming that the supervised specialization of VGGT representations is necessary to recover high-fidelity co-visibility graphs.

Even without any training, the zero-shot variant achieves 0.50~IoU* on the Gibson pairwise setting—already competitive with several supervised baselines (Tab.~\ref{tab:zeroshot_ablation})—confirming that VGGT's frozen representations carry a meaningful co-visibility signal that supervised fine-tuning fully unlocks.

\begin{table}[t]
\centering
\small
\caption{\textbf{Pairwise co-visibility prediction}. Results on Gibson and HM3D datasets. Baselines results are reported from \cite{chen2025covision}, no human baseline available in this case.}
\label{tab:covis_pairwise}
\begin{tabular}{lcccc}
\toprule
\textbf{Method} &
\multicolumn{2}{c}{\textbf{Gibson}} & 
\multicolumn{2}{c}{\textbf{HM3D}} \\
\cmidrule(lr){2-3} \cmidrule(lr){4-5}
& \textbf{IoU* ($\uparrow$)} & \textbf{AUC ($\uparrow$)} 
& \textbf{IoU* ($\uparrow$)} & \textbf{AUC ($\uparrow$)} \\
\midrule
SuperGlue~\cite{sarlin2020superglue}      & 0.47 & 0.11 & 0.38 & 0.10 \\
SIFT + RANSAC~\cite{article}    & 0.35 & 0.05 & 0.34 & 0.05 \\
\midrule
NetVLAD~\cite{arandjelovic2016netvlad}          & 0.35 & 0.33 & 0.35 & 0.29 \\
\midrule
DUSt3R~\cite{wang2024dust3r}           & 0.54 & 0.54 & 0.40 & 0.40 \\
\midrule
ViT~\cite{dosovitskiy2020image}              & 0.47 & 0.43 & 0.48 & 0.46 \\
VGG~\cite{simonyan2014very}              & 0.35 & 0.30 & 0.34 & 0.21 \\
ResNet18~\cite{he2016deep}         & 0.38 & 0.36 & 0.48 & 0.46 \\
CroCo v2~\cite{weinzaepfel2023croco}         & 0.50 & 0.45 & 0.43 & 0.38 \\
Covis~\cite{chen2025covision}            & 0.56 & 0.54 & 0.53 & 0.51 \\
\midrule
Qwen2.5-VL 72B~\cite{bai2025qwen25vltechnicalreport}   & 0.41 & 0.41 & 0.39 & 0.39 \\
Gemini-2.0-Flash~\cite{google2023gemini} & 0.42 & 0.42 & 0.39 & 0.39 \\
SpatialRGPT~\cite{cheng2024spatialrgpt}      & 0.49 & 0.49 & 0.37 & 0.37 \\
GPT-4o~\cite{openai2023gpt4}           & 0.58 & 0.58 & 0.54 & 0.54 \\
\midrule
\rowcolor{gray!10} Co-VGGT (Zero-shot) & 0.50 & 0.31 & 0.46 & 0.31 \\
\rowcolor{gray!10} \textbf{Co-VGGT (Ours)} 
& \textbf{0.85} 
& \textbf{0.78} 
& \textbf{0.84} 
& \textbf{0.78} \\
\bottomrule
\end{tabular}

\end{table}

\begin{figure}[t]
  \centering
  \includegraphics[width=\columnwidth]%{viz_att.pdf}
  {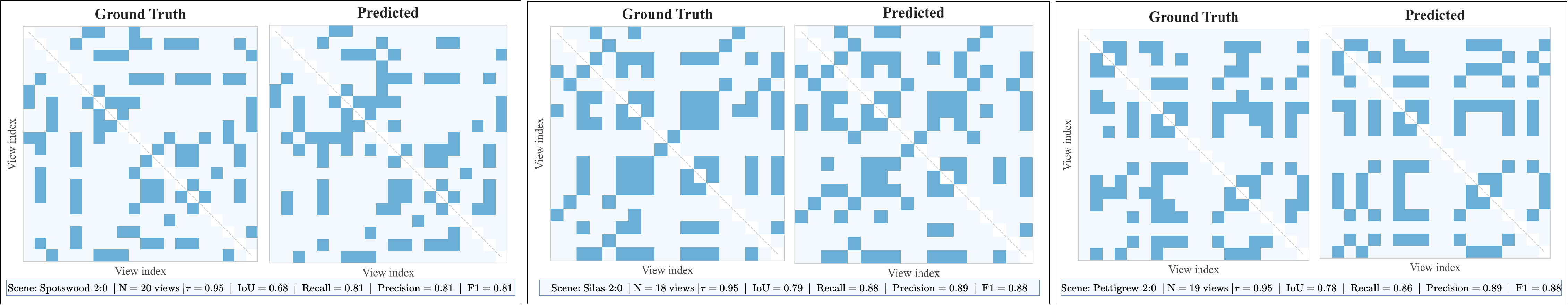}
\caption{\textbf{Co-visibility Scene-Graph Examples (Multiview).} Given $N$ input views, we visualize the ground-truth adjacency matrix (left), the predicted binary graph obtained by thresholding scores at the IoU-optimal $\tau^{*}$. Each matrix is symmetric and entry $(i,j)$ denotes the relation between image $i$ and image $j$.}
 \label{fig:viz_graph}
\end{figure}

\begin{table}[t]
\centering
\small
\caption{\textbf{Multiview co-visibility prediction}. Results on Gibson and HM3D datasets. Baselines and Human results are reported from \cite{chen2025covision}.}
\label{tab:covis_multiview}
\begin{tabular}{lcccc}
\toprule
\textbf{Method} &
\multicolumn{2}{c}{\textbf{Gibson}} &
\multicolumn{2}{c}{\textbf{HM3D}} \\
\cmidrule(lr){2-3} \cmidrule(lr){4-5}
& \textbf{IoU* ($\uparrow$)} & \textbf{AUC ($\uparrow$)}
& \textbf{IoU* ($\uparrow$)} & \textbf{AUC ($\uparrow$)} \\
\midrule
\rowcolor{gray!10}
Human Annotation~\cite{chen2025covision} & 0.72 & 0.72 & -- & -- \\
\midrule
NetVLAD~\cite{arandjelovic2016netvlad}          & 0.38 & 0.32 & 0.42 & 0.39 \\
MV-DUSt3R+~\cite{tang2024mvdust3r}       & 0.56 & 0.56 & 0.45 & 0.45 \\
CroCo v2~\cite{weinzaepfel2023croco}         & 0.48 & 0.41 & 0.41 & 0.37 \\
\midrule
Covis~\cite{chen2025covision}            & 0.59 & 0.57 & 0.57 & 0.56 \\
Covis-freeze~\cite{chen2025covision}     & 0.61 & 0.57 & 0.58 & 0.56 \\
\midrule
GPT-4o~\cite{openai2023gpt4}           & 0.63 & 0.63 & 0.59 & 0.59 \\
\midrule
\rowcolor{gray!10} Co-VGGT (Zero-shot) & 0.37 & 0.30 & 0.36 & 0.30 \\
\rowcolor{gray!10} \textbf{Co-VGGT (Ours)} 
& \textbf{0.74} & \textbf{0.73} 
& \textbf{0.76} & \textbf{0.74} \\
\bottomrule
\end{tabular}
\end{table}

\begin{figure}[t]
  \centering
  \includegraphics[width=\columnwidth]{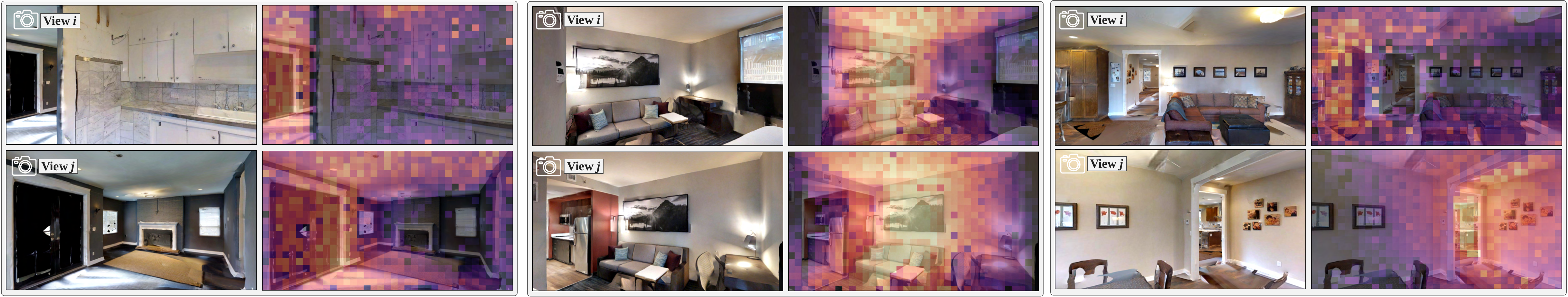}
\caption{\textbf{Cross-view Similarity Matrix Examples.} Examples of \approach\ attention token similarity over pair of co-visibile images. (left) input RGB image, (right) the attention output mask. Features are extracted from signal belonging to layer 17 of the Pairwise evaluation.}
 \label{fig:viz_att}
\end{figure}

\subsection{Pair Aggregation}
\label{sec:pair_aggregation}

Tab.~\ref{tab:ablation_agg} studies how the choice of symmetric pair features impacts performance. 
Using only raw embeddings (simple) is weakest, while adding multiplicative interactions ($e_i\odot e_j$) improves results (product). 
Including absolute differences ($|e_i-e_j|$) provides the largest single gain, suggesting that relative feature offsets capture key overlap cues. 
Our full aggregation $([e_i,e_j], |e_i-e_j|, e_i\odot e_j)$ combined with the layer-wise MoE yields the best IoU*/AUC, outperforming the same feature set under the standard (non-MoE) aggregation baseline.

Notably, the \texttt{absolute} variant matches Co-VGGT in IoU* (0.74) but lags in AUC (0.71 vs.\ 0.73), suggesting that the  element-wise product $\mathbf{e}_i \odot \mathbf{e}_j$ contributes  primarily to probability calibration rather than threshold-optimal 
classification: a meaningful distinction when predicted scores are used as continuous edge weights in downstream visibility graphs.

\begin{table*}[t]
\centering
\scriptsize
\setlength{\tabcolsep}{1.5pt}
\renewcommand{\arraystretch}{1.15}
\caption{\textbf{Performance across difficulty levels.}
Left: edge-level difficulty defined by pairwise image overlap (Easy $\ge 50\%$, Med. $10$--$50\%$, Hard $<10\%$).
Right: graph-level difficulty defined by scene sparsity via average overlap (Easy $\ge 10\%$, Med. $4$--$10\%$, Hard $<4\%$).
Metrics report \emph{Graph IoU} at the best threshold.}
\label{tab:difficulty_levels}
\begin{subtable}[t]{0.49\textwidth}
\centering
\caption{\textbf{Image overlap} (edge-level).}
\label{tab:difficulty_overlap}
\begin{tabular}{lcccc}
\toprule
\textbf{Method} & \textbf{Easy} & \textbf{Med.} & \textbf{Hard} & \textbf{Avg.} \\
\midrule
GPT-4o~\cite{openai2023gpt4}                         & 0.97 & 0.92 & 0.34 & 0.63 \\
Gemini-2.0-Flash~\cite{google2023gemini}             & \textbf{1.00} & \textbf{0.99} & 0.14 & 0.42 \\
Qwen2.5-VL 72B~\cite{bai2025qwen25vltechnicalreport} & \textbf{1.00} & \textbf{0.99} & 0.14 & 0.41 \\
SpatialRGPT~\cite{cheng2024spatialrgpt}              & \textbf{1.00} & \textbf{0.99} & 0.13 & 0.35 \\
Covis~\cite{chen2025covision}                        & 0.99 & 0.88 & 0.24 & 0.59 \\
Covis (freeze)~\cite{chen2025covision}               & \textbf{1.00} & 0.89 & 0.30 & 0.61 \\
\midrule
\rowcolor{gray!10}
Co-VGGT (Multi)               & 0.80 & 0.88 & 0.70 & 0.74 \\
\rowcolor{gray!10}
Co-VGGT (Pair)                    & \textbf{1.00} & 0.93 & \textbf{0.84} & \textbf{0.85} \\
\bottomrule
\end{tabular}
\end{subtable}
\hfill
\begin{subtable}[t]{0.49\textwidth}
\centering
\caption{\textbf{Scene sparsity} (graph-level).}
\label{tab:difficulty_sparsity}
\begin{tabular}{lcccc}
\toprule
\textbf{Method} & \textbf{Easy} & \textbf{Med.} & \textbf{Hard} & \textbf{Avg.} \\
\midrule
GPT-4o~\cite{openai2023gpt4}                         & 0.83 & 0.64 & 0.57 & 0.63 \\
Gemini-2.0-Flash~\cite{google2023gemini}             & 0.75 & 0.39 & 0.28 & 0.42 \\
Qwen2.5-VL 72B~\cite{bai2025qwen25vltechnicalreport} & 0.77 & 0.40 & 0.28 & 0.41 \\
SpatialRGPT~\cite{cheng2024spatialrgpt}              & 0.72 & 0.38 & 0.26 & 0.35 \\
Covis~\cite{chen2025covision}                        & 0.80 & 0.63 & 0.52 & 0.59 \\
Covis (freeze)~\cite{chen2025covision}               & 0.81 & 0.65 & 0.54 & 0.61 \\
\midrule
\rowcolor{gray!10}
Co-VGGT (Multi)                   & 0.99 & 0.92 & 0.73 & 0.74 \\
\rowcolor{gray!10}
Co-VGGT (Pair)                  & \textbf{1.00} & \textbf{0.97} & \textbf{0.84} & \textbf{0.85} \\
\bottomrule
\end{tabular}
\end{subtable}

\end{table*}

\begin{table}[t]
\centering
\scriptsize
\caption{\textbf{Aggregator ablation.} 
All aggregators use MoE except the standard aggregation baseline. 
“Standard” denotes concatenation of $([e_i, e_j], |e_i-e_j|, e_i\odot e_j)$.}
\label{tab:ablation_agg}
\setlength{\tabcolsep}{5.pt}
\renewcommand{\arraystretch}{1.05}
\begin{tabular}{lccc c cc cc}
\toprule
& \multicolumn{3}{c}{Pair feature set} & MoE 
& \multicolumn{2}{c}{\textbf{Gibson}} 
& \multicolumn{2}{c}{\textbf{HM3D}} \\
\cmidrule(lr){2-4}
\cmidrule(lr){6-7}
\cmidrule(lr){8-9}
\textbf{Pair Aggregator} &
$|e_i-e_j|$ & $e_i\odot e_j$ & $[e_i,e_j]$ &
w/ or w/o &
IoU* & AUC &
IoU* & AUC \\
\midrule

simple   
& \xmark & \xmark & \cmark & \cmark 
& 0.67 & 0.70 
& 0.69 & 0.70 \\

product  
& \xmark & \cmark & \cmark & \cmark 
& 0.71 & 0.70 
& 0.72 & 0.71 \\

absolute 
& \cmark & \xmark & \cmark & \cmark 
& 0.74 & 0.71 
& 0.73 & 0.72 \\

standard  
& \cmark & \cmark & \cmark & \xmark 
& 0.69 & 0.68 
& 0.72 & 0.70 \\

\rowcolor{gray!10}
\textbf{\approach} (Ours) 
& \cmark & \cmark & \cmark & \cmark 
& \textbf{0.74} & \textbf{0.73} 
& \textbf{0.76} & \textbf{0.74} \\

\bottomrule
\end{tabular}
\end{table}

\subsection{Hierarchical Information}
\begin{figure}[t]
  \centering
  \includegraphics[width=\columnwidth]{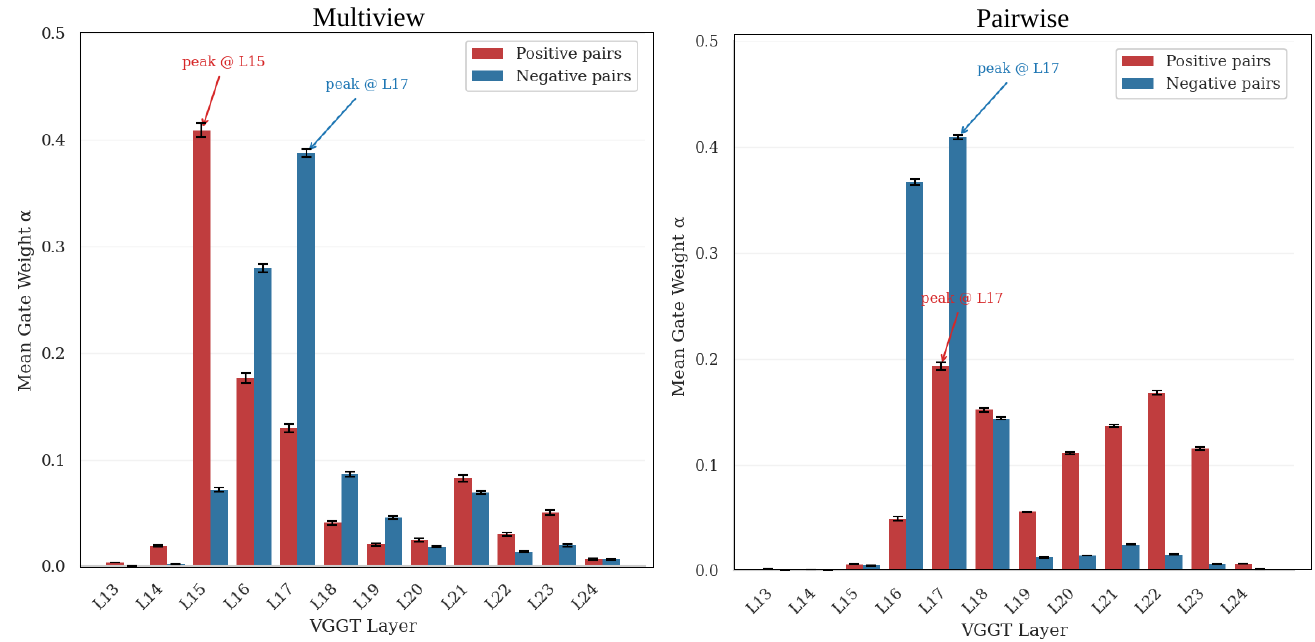}
\caption{\textbf{MoE Gating weights.} Average $\alpha$ parameter per-layer vs. layer id. on the multiview (left) and pairwise (right) task. We observe that specific layers are decisive for the final co-visibility analysis. }
  \label{fig:layer_weights}
\end{figure}

As shown in Fig.~\ref{fig:layer_weights}, co-visibility reasoning is concentrated in VGGT's late layers. 
Gate mass is essentially zero for layers 0–12, confirming that early representations encode geometric primitives rather than overlap cues.
Among late layers, we observe a strong class asymmetry. Negative pairs consistently route to L17 (peak $\approx$0.37–0.41) with low gating entropy across both pairwise and multiview settings, making L17 a negative anchor that confidently rejects geometrically disjoint pairs. 

% This is supported by attention maps (Fig.~\ref{fig:attention_maps} of the supplementary material): for non-co-visible pairs, the first view yields focused, structured attention, while the second produces diffuse, unanchored responses — a signature of failed cross-image correspondence at a geometrically mature layer.
This is supported by attention maps (see supplementary material): for non-co-visible pairs, the first view yields focused, structured attention, while the second produces diffuse, unanchored responses — a signature of failed cross-image correspondence at a geometrically mature layer.
Positive pairs, by contrast, show higher gating entropy and a setting-dependent preference: multiview routing sharpens to L15 (peak $\approx$0.41), while pairwise spreads across L18–L23 (peak $\approx$0.19–0.23), reflecting greater uncertainty about which layer carries the overlap signal when no auxiliary views are available.

This structure directly explains the calibration gap reported below: multiview exhibits sharp, consistent routing but noisier embeddings from compressing multiview context into fixed-size vectors; pairwise produces cleaner embeddings and better calibration but has flatter routing due to the absence of additional scene context.

\subsection{Calibration Measures}

\begin{figure}[t]
  \centering
  \includegraphics[width=\columnwidth]{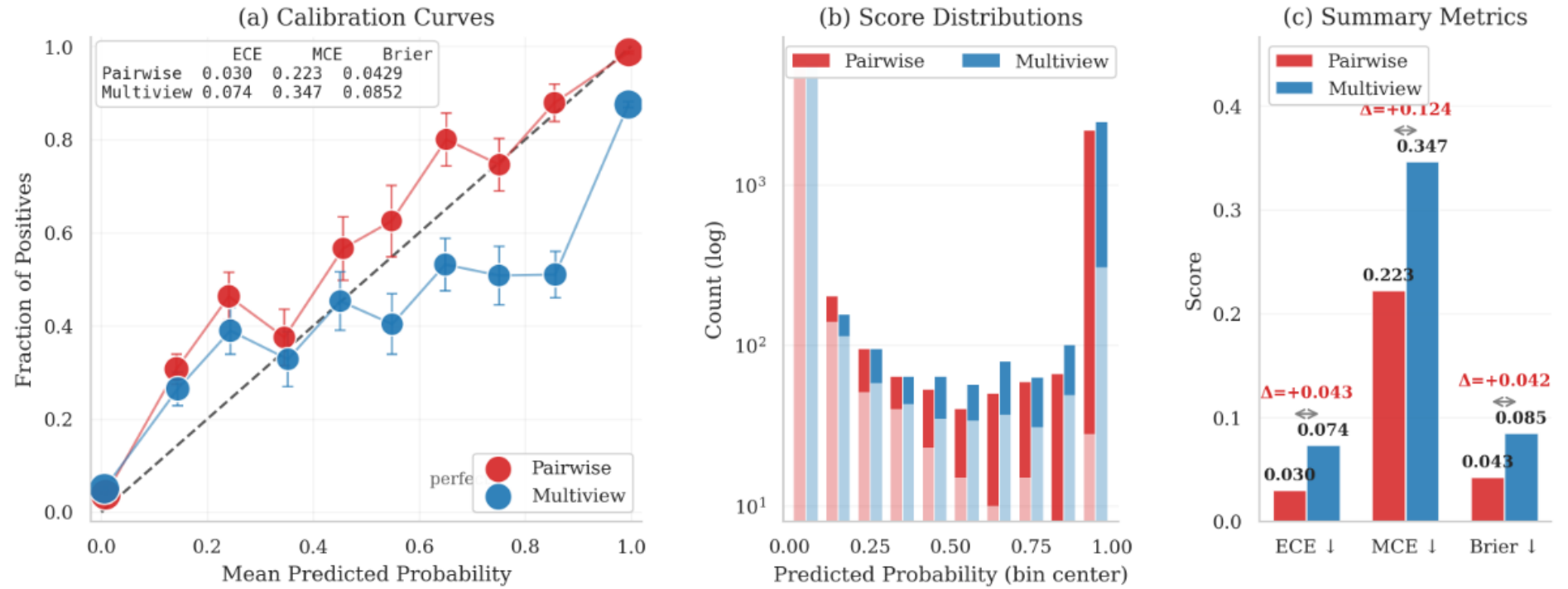}
  \caption{\textbf{Calibration Measures.} Calibration plot (left), score distribution (center) and summary metrics (right) on Gibson validation for pairwise and multiview tasks.}
  \label{fig:calibration}
\end{figure}

In Fig.~\ref{fig:calibration}, we assess probability calibration on Gibson val using Expected Calibration Error (ECE), Maximum Calibration Error (MCE), and
Brier score.
In the pairwise setting, \approach\ is well calibrated (ECE\,$=$\,0.030, Brier\,$=$\,0.043): predicted scores behave as empirical frequencies (e.g., $p{\approx}0.7$ implies ${\sim}70\%$ true positives), making them directly usable as edge weights for downstream filtering  keeping $p{>}0.7$ for matching and discarding $p{<}0.2$ (to suppress spurious constraints) without any post-hoc correction.
In multiview, calibration degrades (ECE\,$=$\,0.074, Brier\,$=$\,0.085), consistent with the noisier embeddings produced by compressing multiview context into fixed-size vectors, as discussed in Sec.~\ref{sec:layer_weights}.
Both settings exhibit high MCE (0.223 pairwise, 0.347 multiview); however, this is driven by sparsely populated mid-confidence bins, since over 90\% of predictions fall in $[0,\,0.1)$ or $[0.9,\,1]$.
%In the pairwise setting Co-VGGT is well calibrated (ECE = 0.030): scores behave as empirical frequencies (p$\sim$0.7 $\rightarrow$ $\sim$70\% true positives), usable directly as edge weights — keep p>0.7, discard p<0.2 — without post-hoc correction. Multiview calibration degrades (Fig.~\ref{fig:calibration}), consistent with its noisier compressed embeddings. The high MCE in both settings is an artifact of sparsely populated mid-confidence bins (>90\% of predictions fall in [0,0.1) or [0.9,1]), so MCE is therefore unrepresentative of typical operating thresholds, where the model is strongly decisive and well-behaved.
Furthermore, we added a proof-of-concept COLMAP experiment against different methods to show this downstream capability (see Supplementary material).

\subsection{Layer Ablation}
\label{sec:layer_weights}
Tab.~\ref{tab:ablation_layers_mode} shows that restricting the MoE expert pool to the last 12 layers preserves nearly full performance (0.74/0.72 IoU*/AUC on Gibson), while using fewer layers causes consistent degradation. 
This is directly corroborated by the gating analysis (Fig.~\ref{fig:layer_weights}), where layers 0–12 receive near-zero average weight across both classes, confirming that the model has learned to ignore early layers. 
Since VGGT inference dominates runtime (50ms/150ms per 2/10 images on an H100) and the MoE head is lightweight, retaining all layers adds negligible overhead — we therefore do so in our final model. Together, these findings provide task-grounded empirical evidence for the hierarchical geometric reasoning hypothesized in~\cite{wang2025vggt}: early layers carry no discriminative signal for co-visibility, while late layers encode the high-level judgment of accepting or rejecting shared surface support, with positive and negative pairs peaking at distinct layers (L15 and L17, respectively).
For completeness, we evaluate our approach using only the first 12 layers of the VGGT backbone (last row of Tab.~\ref{tab:ablation_layers_mode}). The resulting performance drops substantially below the state of the art, confirming that early layers encode more generic and diverse representations that are less informative for co-visibility reasoning.

\subsection{Cross-domain transfer}
To probe whether the learned co-visibility signal is tied to a single training distribution, we evaluate Co-VGGT trained on one Co-VisiON environment and tested zero-shot on the other (Tab.~\ref{tab:ablation_layers_mode}). Transferring across Gibson-HM3D costs at most 0.04 IoU* and 0.03 AUC in both pairwise and multiview settings, while still exceeding the in-domain state of the art on the target dataset. Combined with Co-VisiON's scene-disjoint train/val protocol, this indicates the extracted signal does not overfit to a single environment, though both datasets remain indoor Habitat-rendered scenes and broader cross-domain generalization is left to future work.

\begin{table}[t]
\centering
\scriptsize
\setlength{\tabcolsep}{2pt}
\caption{\textbf{Ablation and cross-domain transfer results.}
Left: ablation on the number of VGGT expert-pool layers used in the MoE head, evaluated on Gibson.
Right: cross-domain transfer between Gibson and HM3D.}
\label{tab:ablation_layers_mode}

\begin{minipage}[t]{0.49\linewidth}
\centering
\resizebox{\linewidth}{!}{%
\begin{tabular}{lcccc}
\toprule
\multicolumn{5}{c}{\textbf{Layer-count ablation}} \\
\midrule
& \multicolumn{2}{c}{\textbf{Multiview}}
& \multicolumn{2}{c}{\textbf{Pairwise}} \\
\cmidrule(lr){2-3} \cmidrule(lr){4-5}
\textbf{MoE Range}
& \textbf{IoU}$^\star \uparrow$ & \textbf{AUC}$\uparrow$
& \textbf{IoU}$^\star \uparrow$ & \textbf{AUC}$\uparrow$ \\
\midrule
\rowcolor{gray!10}
$[1,24]$ (Ours) & \textbf{0.74} & \textbf{0.73} & \textbf{0.84} & \textbf{0.78} \\
$[14,24]$       & \textbf{0.74} & 0.72          & \textbf{0.84} & 0.77 \\
$[19,24]$       & 0.71          & 0.70          & 0.83          & 0.75 \\
$[21,24]$       & 0.70          & 0.67          & 0.83          & 0.75 \\
$[24]$          & 0.69          & 0.66          & 0.82          & 0.74 \\
$[1,12]$        & 0.53          & 0.50          & 0.30          & 0.11 \\
\bottomrule
\end{tabular}%
}
\end{minipage}
\hfill
\begin{minipage}[t]{0.49\linewidth}
\centering
\resizebox{\linewidth}{!}{%
\begin{tabular}{llcccc}
\toprule
\multicolumn{6}{c}{\textbf{Cross-domain transfer}} \\
\midrule
\textbf{Train} & \textbf{Test}
& \multicolumn{2}{c}{\textbf{Pairwise}}
& \multicolumn{2}{c}{\textbf{Multiview}} \\
\cmidrule(lr){3-4} \cmidrule(lr){5-6}
&
& \textbf{IoU}$^\star \uparrow$ & \textbf{AUC}$\uparrow$
& \textbf{IoU}$^\star \uparrow$ & \textbf{AUC}$\uparrow$ \\
\midrule
\rowcolor{gray!10}
Gibson & Gibson & \textbf{0.85} & \textbf{0.78} & \textbf{0.74} & \textbf{0.73} \\
\rowcolor{gray!10}
HM3D   & HM3D   & \textbf{0.84} & \textbf{0.78} & \textbf{0.76} & \textbf{0.74} \\
\midrule
Gibson & HM3D   & 0.81 & 0.75 & 0.72 & 0.71 \\
HM3D   & Gibson & 0.83 & 0.76 & 0.72 & 0.72 \\
\bottomrule
\end{tabular}%
}
\end{minipage}

\end{table}

\subsection{Zero-Shot Ablation}
As shown in Tab.~\ref{tab:covis_pairwise}, the zero-shot performance of \approach\ is unexpectedly strong, indicating that the backbone already encodes a substantial signal for identifying co-visible pairs. 
%In this regime, simple feature extraction is sufficient to obtain competitive results.

To further investigate this behavior, Tab.~\ref{tab:zeroshot_ablation} presents a zero-shot ablation where we vary the subset of VGGT backbone layers used for similarity computation and subsequent IoU and AUC evaluation (analogous to Tab.~\ref{tab:ablation_layers_mode}). Performance remains within a relatively narrow range across different subsets. 
However, across all tasks and datasets, the strongest results are consistently achieved when using embeddings from the last five layers.
%This suggests that the optimal zero-shot configuration relies on features from the final five backbone layers. 
%Additional analyses and extended experiments are provided in Sec.~\ref{sec:supp_zeroshot} of the Supplementary material.
Additional analyzes and extended experiments are provided in Sec.~A.1 of the Supplementary material.

\begin{table}[t]
\centering
\caption{\textbf{Zero-shot layer ablation.}
Ablation on the number of VGGT output feature layers used for cosine similarity scoring in zero-shot evaluation, reported on both Gibson and HM3D datasets for Multiview and Pairwise tasks.}
\label{tab:zeroshot_ablation}
\scriptsize
\setlength{\tabcolsep}{3pt}
\renewcommand{\arraystretch}{1.05}

\begin{tabular}{lcccccccc}
\toprule
& \multicolumn{4}{c}{\textbf{Gibson}} 
& \multicolumn{4}{c}{\textbf{HM3D}} \\
\cmidrule(lr){2-5} \cmidrule(lr){6-9}
\textbf{Layer Range}
& \multicolumn{2}{c}{\textbf{Multiview}}
& \multicolumn{2}{c}{\textbf{Pairwise}}
& \multicolumn{2}{c}{\textbf{Multiview}}
& \multicolumn{2}{c}{\textbf{Pairwise}} \\
\cmidrule(lr){2-3} \cmidrule(lr){4-5}
\cmidrule(lr){6-7} \cmidrule(lr){8-9}
& \textbf{IoU*$\scriptsize(\uparrow)$} & \textbf{AUC$\scriptsize(\uparrow)$}
& \textbf{IoU*$\scriptsize(\uparrow)$} & \textbf{AUC$\scriptsize(\uparrow)$}
& \textbf{IoU*$\scriptsize(\uparrow)$} & \textbf{AUC$\scriptsize(\uparrow)$}
& \textbf{IoU*$\scriptsize(\uparrow)$} & \textbf{AUC$\scriptsize(\uparrow)$} \\
\midrule

\rowcolor{gray!10}
$[1,24]$ (Ours)
& 0.37 & 0.30 & 0.50 & 0.31
& 0.36 & \textbf{0.30} & 0.46 & 0.31 \\

%$[4,24]$
%& 0.37 & 0.30 & 0.50 & 0.32
%& 0.36 & \textbf{0.30} & 0.46 & 0.31 \\

$[9,24]$
& 0.37 & 0.30 & 0.50 & 0.32
& 0.36 & \textbf{0.30} & 0.46 & 0.31 \\

%$[14,24]$
%& 0.37 & 0.30 & 0.51 & 0.33
%& 0.36 & \textbf{0.30} & 0.46 & 0.31 \\

%$[16,24]$
%& 0.36 & \textbf{0.31} & 0.51 & 0.34
%& 0.36 & \textbf{0.30} & 0.46 & 0.31 \\

$[19,24]$
& \textbf{0.38} & \textbf{0.31} & \textbf{0.52} & \textbf{0.36}
& \textbf{0.37} & \textbf{0.30} & \textbf{0.47} & \textbf{0.33} \\

$[21,24]$
& 0.36 & 0.30 & 0.49 & 0.34
& 0.36 & \textbf{0.30} & 0.45 & 0.30 \\

%$[24]$
%& 0.34 & 0.30 & 0.41 & 0.29
%& 0.34 & 0.29 & 0.37 & 0.27 \\

\bottomrule
\end{tabular}
\end{table}

\section{Limitations}
\label{sec:limitations}
While Co-VGGT achieves strong results on the Co-VisiON benchmark, some limitations remain. 
We frame co-visibility prediction as a binary signal since the focus of this work is primarily on interpretability and understanding the overlap signal already present in VGGT, given that a simple co-visibility estimator already achieves strong performance.
Nevertheless, we also conducted an auxiliary experiment that predicts, via a regression head, a graded co-visibility strength (see Supplementary material). 
We do not further develop this direction here; instead, we leave it as future work, where richer overlap estimates could provide more informative geometric constraints for downstream SfM and SLAM pipelines.

Moreover, the multiview setting exposes a structural limitation of our architecture: compressing multiview context into fixed-dimensional per-view embeddings introduces noise, and the model scores each pair independently without enforcing global graph consistency — leading to a performance gap relative to the pairwise setting. 
Further failure examples and analysis are provided in the Supplementary material.

Finally, the mechanistic interpretation of layer-wise gating — while empirically grounded — remains correlational: we identify \emph{which} layers are decisive, but not entirely \emph{why} specific geometric abstractions emerge at those depths in VGGT's training.
However, we argue that the emergent behavior of VGGT in 3D vision may mirror that of LLMs in NLP; consequently, several insights and solutions could potentially be transferred from that domain.

\section{Conclusion}
\label{sec:conclusion}

We presented Co-VGGT, a lightweight co-visibility predictor on frozen VGGT features, with well-calibrated pairwise probabilities (ECE=0.030) that are usable directly as visibility-graph edge weights. Our layer-wise mixture-of-experts head reveals that co-visibility reasoning
emerges exclusively in VGGT's late transformer blocks: L17 acts as a negative anchor that confidently rejects non-overlapping pairs, while positive pairs rely on a broader set of late layers depending on the available context. This provides task-grounded evidence for hierarchical geometric reasoning in a geometry-grounded foundation model, mirroring the layer-specialization behavior observed in large language models.

These findings suggest that geometry-grounded foundation models encode spatial priors that are richer and more structured than previously understood — and that these priors can be efficiently distilled for downstream tasks without modifying the backbone.
Future work will pursue four directions: (i) moving beyond binary prediction toward continuous overlap ratio estimation to supply richer geometric constraints for SfM and SLAM pipelines; (ii) replacing the current per-pair multiview loop with a unified aggregation mechanism — such as set transformers or token-level routing — to enforce global graph consistency and close the pairwise–multiview calibration gap; and (iii) integrating Co-VGGT within embodied mapping systems to enable uncertainty-aware loop closure and next-best-view planning grounded in geometric connectivity rather than appearance alone.
(iv) Investigate whether interpretability findings from NLP transfer to geometric foundation models, enabling principled interpretability methods for this domain.
%Future work includes continuous overlap-ratio estimation for richer SfM/SLAM constraints, a unified multiview aggregation mechanism (e.g., set transformers) to enforce graph consistency and close the calibration gap, and integration into embodied mapping for uncertainty-aware loop closure and next-best-view planning.

%\clearpage  % TODO FINAL: This \clearpage needs to be removed from both review and camera-ready versions.

\section*{Acknowledgements}
We thank Fondazione Bruno Kessler (FBK) and the University of Padua, Department of Mathematics "Tullio Levi-Civita", for providing the computational resources used in this work. TC and LS were supported by the PNRR project Future Artificial Intelligence Research (FAIR, PE00000013) under the NRRP MUR program, funded by NextGenerationEU.

%Please insert your acknowledgments here.

\bibliographystyle{splncs04}
\bibliography{main}

%\clearpage

% --- Supplementary numbering ---
\setcounter{figure}{0}
\setcounter{table}{0}
\setcounter{page}{1}          % <-- add this
\renewcommand{\thefigure}{S\arabic{figure}}
\renewcommand{\thetable}{S\arabic{table}}

\title{What VGGT Knows About Overlap: Probing Geometric Foundation Models for Co-Visibility}
\titlerunning{What VGGT Knows About Overlap}
\authorrunning{F. Ziliotto et al.}

\author{
Filippo Ziliotto\inst{1,2} \and
Luciano Serafini\inst{2} \and
Lamberto Ballan\inst{1} \and
Tommaso Campari\inst{2}
}

\institute{
University of Padova \and
Fondazione Bruno Kessler (FBK)
}

\maketitle

\appendix
\section{Supplementary Material}
\label{sec:supplementary}

We provide supplementary material to complement the main paper. Sec.~\ref{sec:supp_zeroshot} reports extended zero-shot ablations across layer subsets and datasets. Sec.~\ref{sec:supp_mv_pairwise} evaluates the multiview-trained model under pairwise inference. Sec.~\ref{sec:supp_metrics} defines the full set of evaluation metrics used throughout. Sec.~\ref{sec:supp_failures} presents qualitative failure analysis and scene-graph visualizations. Sec.~\ref{sec:supp_finetune} ablates end-to-end VGGT fine-tuning as an upper-bound reference.
Sec.~\ref{supp:strenght_pred} explains the addition of the regression head to the current \approach\ approach.

\subsection{Zero-shot Ablation}
\label{sec:supp_zeroshot}

As discussed in Sec.~4, we report additional zero-shot ablations in Tab.~\ref{tab:zeroshot_ablation_full}, where we carefully define the layer ranges used to construct the MoE heads. Notably, the best performance consistently emerges from the [19,24] range across all settings and datasets.

We do not investigate this behavior further. It may be partially influenced by the specific strategy adopted to extract representations in the zero-shot regime. Consequently, we do not claim that this choice is optimal, but rather present it as a reasonable and reproducible approach to address the zero-shot setting.

We further observe that the results in Tab.~\ref{tab:zeroshot_ablation_full} exhibit only minor variations across different layer ranges, which may partially stem from randomness induced by different statistical seeds.

\begin{table}[th]
\centering
\caption{\textbf{Zero-shot layer ablation.}
Ablation on the number of VGGT output feature layers used for cosine similarity scoring in zero-shot evaluation, reported on both Gibson and HM3D datasets for Multiview and Pairwise tasks.}
\label{tab:zeroshot_ablation_full}
\scriptsize
\setlength{\tabcolsep}{3pt}
\renewcommand{\arraystretch}{1.05}

\begin{tabular}{lcccccccc}
\toprule
& \multicolumn{4}{c}{\textbf{Gibson}} 
& \multicolumn{4}{c}{\textbf{HM3D}} \\
\cmidrule(lr){2-5} \cmidrule(lr){6-9}
\textbf{Layer Range}
& \multicolumn{2}{c}{\textbf{Multiview}}
& \multicolumn{2}{c}{\textbf{Pairwise}}
& \multicolumn{2}{c}{\textbf{Multiview}}
& \multicolumn{2}{c}{\textbf{Pairwise}} \\
\cmidrule(lr){2-3} \cmidrule(lr){4-5}
\cmidrule(lr){6-7} \cmidrule(lr){8-9}
& \textbf{IoU*$\scriptsize(\uparrow)$} & \textbf{AUC$\scriptsize(\uparrow)$}
& \textbf{IoU*$\scriptsize(\uparrow)$} & \textbf{AUC$\scriptsize(\uparrow)$}
& \textbf{IoU*$\scriptsize(\uparrow)$} & \textbf{AUC$\scriptsize(\uparrow)$}
& \textbf{IoU*$\scriptsize(\uparrow)$} & \textbf{AUC$\scriptsize(\uparrow)$} \\
\midrule

\rowcolor{gray!10}
$[1,24]$ \textbf{(Ours)}
& 0.37 & 0.30 & 0.50 & 0.31
& 0.36 & \textbf{0.30} & 0.46 & 0.31 \\

$[4,24]$
& 0.37 & 0.30 & 0.50 & 0.32
& 0.36 & \textbf{0.30} & 0.46 & 0.31 \\

$[9,24]$
& 0.37 & 0.30 & 0.50 & 0.32
& 0.36 & \textbf{0.30} & 0.46 & 0.31 \\

$[14,24]$
& 0.37 & 0.30 & 0.51 & 0.33
& 0.36 & \textbf{0.30} & 0.46 & 0.31 \\

$[16,24]$
& 0.36 & \textbf{0.31} & 0.51 & 0.34
& 0.36 & \textbf{0.30} & 0.46 & 0.31 \\

$[19,24]$
& \textbf{0.38} & \textbf{0.31} & \textbf{0.52} & \textbf{0.36}
& \textbf{0.37} & \textbf{0.30} & \textbf{0.47} & \textbf{0.33} \\

$[21,24]$
& 0.36 & 0.30 & 0.49 & 0.34
& 0.36 & \textbf{0.30} & 0.45 & 0.30 \\

$[24]$
& 0.34 & 0.30 & 0.41 & 0.29
& 0.34 & 0.29 & 0.37 & 0.27 \\

\bottomrule
\end{tabular}

\end{table}

\subsection{Multiview on Pairwise data}
\label{sec:supp_mv_pairwise}

Since pairwise evaluation is a special case of the multiview setting—where the model receives $N$ views of the same scene—we evaluated \approach, trained in the multiview regime, under pairwise evaluation (Tab.~\ref{tab:multiview_on_pairwise}).
We observe a slight performance drop compared to standard multiview training and testing. We attribute this to a shift in the MoE dynamics (Fig.~\ref{fig:layer_weights}), which alters the layer weighting: some layers are required to produce outputs outside their primary specialization, leading to suboptimal performance.

This suggests an interesting research direction, likely tied to designing architectures that natively support multiview reasoning rather than iterating over pairs, as currently done in \approach\ —a known limitation. We encourage further investigation of this behavior.

\begin{table}[t]
\centering
\small
\caption{\textbf{Multiview on Pairwise data.} We evaluate \approach, trained on the multiview setting, on the pairwise dataset. }
\label{tab:multiview_on_pairwise}
\begin{tabular}{lcccc}
\toprule
\textbf{Method} &
\multicolumn{2}{c}{\textbf{Gibson}} & 
\multicolumn{2}{c}{\textbf{HM3D}} \\
\cmidrule(lr){2-3} \cmidrule(lr){4-5}
& \textbf{IoU* ($\uparrow$)} & \textbf{AUC ($\uparrow$)} 
& \textbf{IoU* ($\uparrow$)} & \textbf{AUC ($\uparrow$)} \\
\midrule
 Co-VGGT (Zero-shot) & 0.50 & 0.31 & 0.46 & 0.31 \\
 Co-VGGT (Multiview) & 0.77 & 0.59 & 0.77 & 0.58 \\
\rowcolor{gray!10} \textbf{Co-VGGT (Ours)} 
& \textbf{0.85} 
& \textbf{0.78} 
& \textbf{0.84} 
& \textbf{0.78} \\
\bottomrule
\end{tabular}

\end{table}

\begin{figure}[t]
  \centering
  \includegraphics[width=\columnwidth]{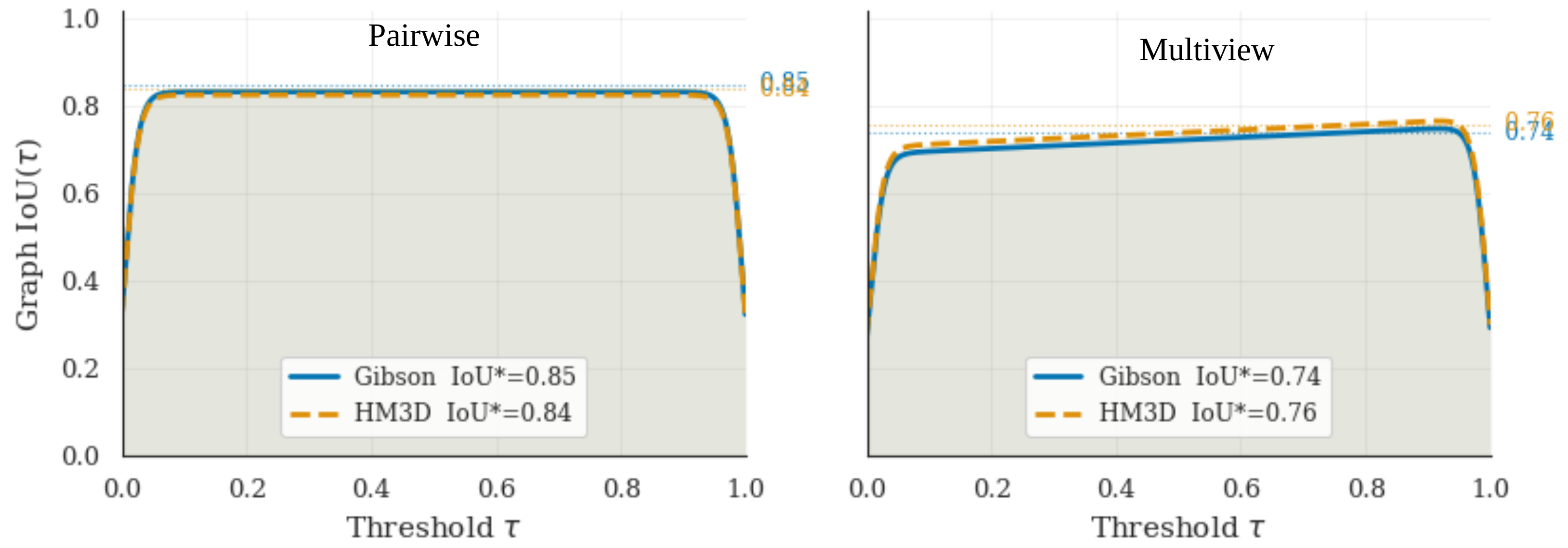}
\caption{\textbf{IoU vs. Thresholds.} Plot of IoU across threshold for different datasets and settings.}
 \label{fig:viz_graph_failures}
\end{figure}

\subsection{Failure Analysis}
\label{sec:supp_failures}
We report several failure cases in Fig.~\ref{fig:viz_graph_failures}. Importantly, the model remains capable of predicting highly overlapping views; failures predominantly arise in medium-to-edge cases. 
In the multiview setting, predicting whether two images overlap is generally easier when the model has access to the remaining $N-2$ views of the scene, particularly those captured in close proximity to the queried pair $(i,j)$. This additional context provides geometric cues that facilitate overlap reasoning. In contrast, when no auxiliary context is available and the pair $(i,j)$ exhibits only minimal overlap, the task becomes significantly more challenging.

As discussed in Sec.~5, a known limitation is that the architecture is primarily designed for pairwise processing. From a practical standpoint, the multiview setting is implemented as a simple loop over all image pairs, rather than through a unified multiview design. We argue that a coherent architecture explicitly tailored for multiview reasoning would likely outperform the current pairwise formulation, as it could more effectively aggregate and leverage information across viewpoints.

\begin{figure}[t]
  \centering
  \includegraphics[width=\columnwidth]%{viz_att.pdf}
  {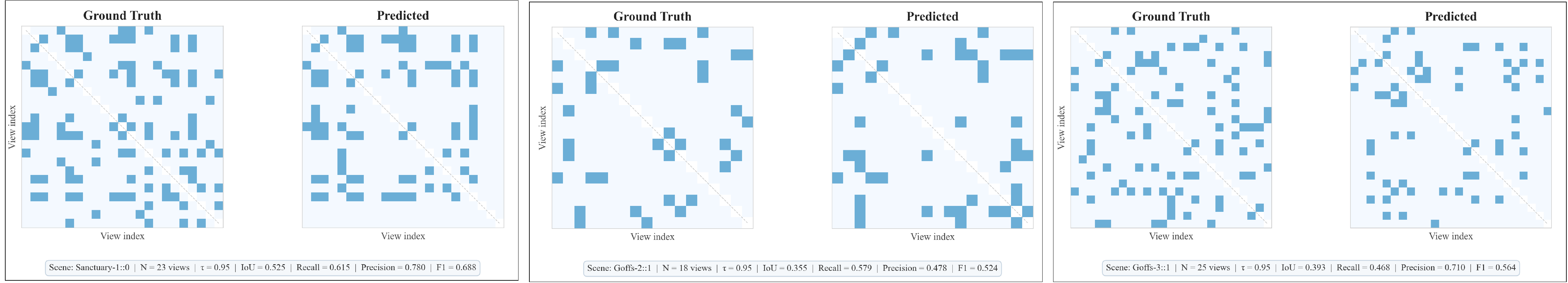}
\caption{\textbf{Co-visibility Scene-Graph Failure Examples (Multiview).} Given $N$ input views, we visualize the ground-truth adjacency matrix (left), the predicted binary graph obtained by thresholding scores at the IoU-optimal $\tau^{*}$. Each matrix is symmetric and entry $(i,j)$ denotes the relation between image $i$ and image $j$.}
 \label{fig:viz_graph_failures}
\end{figure}

\begin{figure}[t]
  \centering
  \includegraphics[width=\columnwidth]%{viz_att.pdf}
  {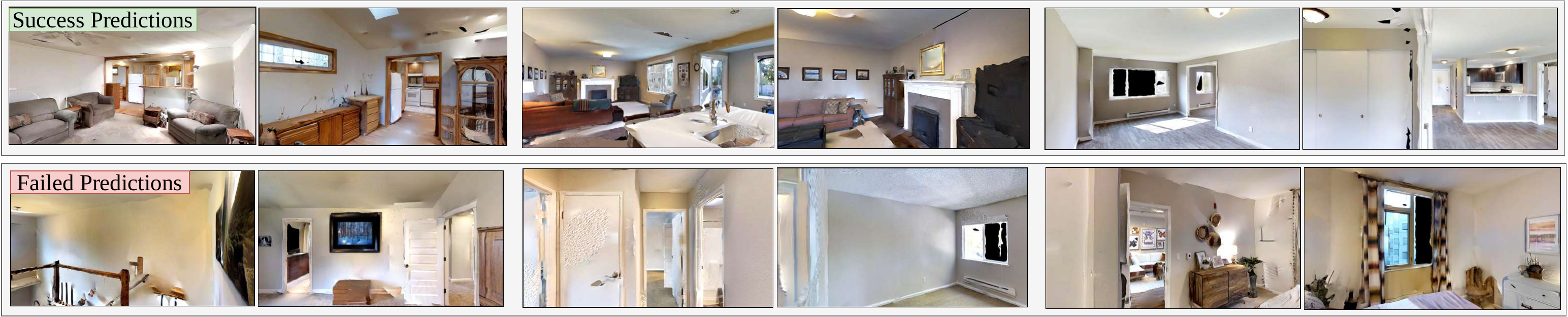}
\caption{\textbf{Qualitative Examples}. Subset of correctly predicted pairs (top) and failed pairs predictions (bottom). Failures are due primarly to the difficulty of the task, since pairs share very minimal overlap between each other.}
 \label{fig:viz_graph_failures}
\end{figure}

\begin{figure}[t]
  \centering
  \includegraphics[width=\columnwidth]%{viz_att.pdf}
  {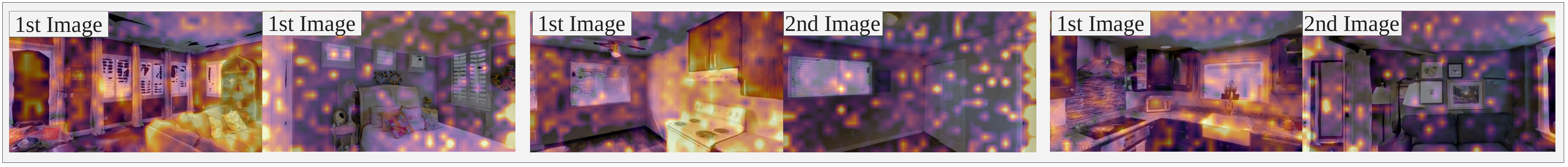}
\caption{\textbf{Attention Maps}. Attention maps extracted from Layer 17 for pairs of non-covisible images. After focusing on objects in the first image, the attention maps for the second image appear sparse and diffuse.}
 \label{fig:attention_maps}
\end{figure}

\subsection{VGGT Finetuning}
\label{sec:supp_finetune}
We further compare \approach\ against a fully fine-tuned VGGT backbone, evaluated both with and without the MoE head (in the latter case, replaced by a standard MLP head). Results in Tab.~\ref{tab:vggt_finetune} show that end-to-end fine-tuning with the MoE head achieves outstanding performance, reaching an IoU/AUC of 0.91/0.88 (last row), effectively saturating the co-visibility task.
Although this demonstrates the strong capacity of the MoE formulation when coupled with full backbone adaptation, improving absolute performance is not the primary objective of this work. Our focus is instead on leveraging the MoE design as a principled and interpretable mechanism to analyze VGGT behavior. For this reason, we do not emphasize these fully fine-tuned results in the main paper.

Moreover, the gap between training only 7.5M parameters and fully fine-tuning a 1.2B-parameter backbone is substantial, both in scale and computational cost. In practice, adapting such a large model is largely impractical for standard SLAM or robotics matching pipelines, where efficiency, deployment constraints, and limited task-specific data typically preclude full end-to-end fine-tuning.

%A similar trend is observed in the multiview setting, where \approach\ with a fully trainable backbone achieves xx/xx in IoU/AUC, respectively.

We note that all fine-tuning experiments were carried out with a $lr=1e^{-5}$ and a $wd=1e^{-5}$, to avoid ``disrupting'' the weights learned before this fine-tuning.

\subsection{Downstream SfM}
\label{sec:colmap}

To test the claim that Co-VGGT scores are usable as visibility-graph weights, we use them as a pair-selection front-end for COLMAP~\cite{schonberger2016structure} on $20$ Gibson scenes. We score all $\binom{N}{2}$ pairs per scene and pass only the top-$30\%$ to feature matching, leaving the rest of the pipeline unchanged,
and compare against exhaustive matching, random selection, and
NetVLAD~\cite{arandjelovic2016netvlad} retrieval under the same budget
(Tab.~\ref{tab:colmap}). Co-VGGT recovers $83\%$ of the registered images and $70\%$ of the sparse points of exhaustive matching while using only $30\%$ of the pairs, outperforming both Random and NetVLAD on coverage. Notably, it also attains the best geometric quality of all methods — including exhaustive — with the highest verification ratio ($0.966$ vs.\ $0.910$) and lowest reprojection error ($0.270$ vs.\ $0.323$) at $2.6\times$ lower runtime. This is a precision effect: discarding low-overlap pairs before verification removes constraints likely to fail, leaving a cleaner pair set. We present this as evidence that Co-VGGT scores are usable for SfM pair selection, not as a
full reconstruction benchmark.

\begin{table}[t]
\centering
\scriptsize
\caption{\textbf{SfM pair-selection Experiment}. Averaged over a small subset of $20$ Gibson scenes.}
\label{tab:colmap}
\setlength{\tabcolsep}{2.2pt}
\begin{tabular}{lcccccc}
\toprule
Selection & Budget & Reg. imgs $\uparrow$ & Points $\uparrow$ & Verif. ratio $\uparrow$ & Reproj. $\downarrow$ & Time (s) $\downarrow$ \\
\midrule
Exhaustive        & 100\% & $12.8 \pm 6.9$ & $3140 \pm 2292$ & $0.910 \pm 0.055$ & $0.323 \pm 0.117$ & $97.0 \pm 60.6$ \\
\midrule
Random top-$k$    & 30\%  & $6.1 \pm 4.3$  & $553 \pm 506$   & $0.917 \pm 0.100$ & $0.301 \pm 0.072$ & $\mathbf{37.3 \pm 19.5}$ \\
NetVLAD top-$k$   & 30\%  & $9.8 \pm 4.8$  & $1945 \pm 1577$ & $0.954 \pm 0.056$ & $0.291 \pm 0.068$ & $40.9 \pm 19.3$ \\
Co-VGGT top-$k$   & 30\%  & $\mathbf{10.6 \pm 4.4}$ & $\mathbf{2086 \pm 1385}$ & $\mathbf{0.966 \pm 0.042}$ & $\mathbf{0.270 \pm 0.058}$ & $37.8 \pm 19.1$ \\
\bottomrule
\end{tabular}
\end{table}

\begin{table}[t]
\centering
\caption{\textbf{VGGT Finetuning Ablation (Gibson).} Comparison between multiview and pairwise evaluation settings. Results are reported on Gibson for both settings. $^{\dagger}$We train the MoE and the full backbone VGGT model.}
\label{tab:vggt_finetune}
\scriptsize
\setlength{\tabcolsep}{4.5pt}
\renewcommand{\arraystretch}{1.05}

\begin{tabular}{lccccccc}
\toprule
\textbf{Method} 
& \textbf{MoE} 
& \textbf{Trainable} 
& \multicolumn{2}{c}{\textbf{Multiview}} 
& \multicolumn{2}{c}{\textbf{Pairwise}}  \\
\cmidrule(lr){4-5} \cmidrule(lr){6-7} 

& \textbf{Head} 
& \textbf{Parameters}
& \textbf{IoU* ($\uparrow$)} 
& \textbf{AUC ($\uparrow$)}
& \textbf{IoU* ($\uparrow$)} 
& \textbf{AUC ($\uparrow$)}\\

\midrule

Co-VGGT (Zero-shot) 
& $\times$ & 0 
& 0.37 & 0.30 
& 0.50 & 0.31 \\

\midrule

VGGT~\cite{wang2025vggt} (Fine-tune)
& $\times$ & 80M 
& 0.72 & 0.69 
& 0.80 & 0.72 \\

VGGT~\cite{wang2025vggt} (Fine-tune) 
& $\times$ & 175M 
& 0.72 & 0.69 
& 0.80 & 0.73 \\

VGGT~\cite{wang2025vggt} (Fine-tune)
& $\times$ & 335M 
& 0.72 & 0.70 
& 0.81 & 0.74 \\

\midrule

VGGT~\cite{wang2025vggt} (Fine-tune) 
& $\times$ & 1.2B 
& 0.72 & 0.71 
& 0.84 & 0.81 \\

Co-VGGT$^{\dagger}$ (Fine-tune) 
& $\checkmark$ & 1.2B 
& 0.76 & 0.72
& \textbf{0.91} & \textbf{0.88} \\

\midrule

\rowcolor{gray!10} Co-VGGT\textbf{ (Ours) }
& $\checkmark$ & 7.5M 
& \textbf{0.74} & \textbf{0.73} 
& 0.85 & 0.78 \\

\bottomrule
\end{tabular}

\end{table}

\subsection{Additional Metrics}
\label{sec:supp_metrics}

\begin{figure}[t]
  \centering
  \includegraphics[width=\columnwidth]{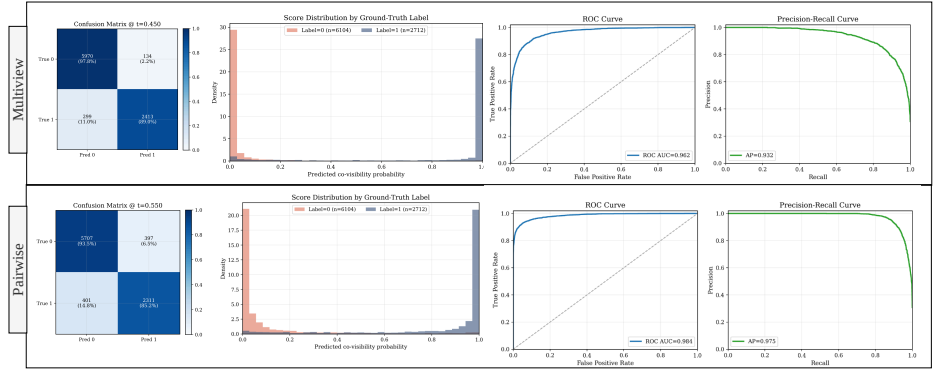}
\caption{\textbf{Additional Metrics}. Other metrics reported in Sec.~\ref{sec:supp_metrics} for Pairwise and Multiview in HM3D dataset. Resulted are reported on a subset of the validation data.}
 \label{fig:additional_metrics}
\end{figure}

Beyond IoU/AUC, we report standard binary classification metrics computed over co-visibility edges (see Fig.~\ref{fig:additional_metrics}).
Accuracy is the fraction of correctly classified view pairs after thresholding predicted co-visibility scores $d_{ij}\in[0,1]$ into labels $\hat{y}_{ij}=\mathbb{I}[d_{ij}\ge\tau]$, i.e., $(\mathrm{TP}+\mathrm{TN})/(\mathrm{TP}+\mathrm{TN}+\mathrm{FP}+\mathrm{FN})$.
Because co-visibility is typically imbalanced (most pairs are non-overlapping), accuracy can be dominated by true negatives and should be interpreted together with ranking-based curves.
The ROC curve plots the True Positive Rate (TPR $=\mathrm{TP}/(\mathrm{TP}+\mathrm{FN})$) against the False Positive Rate (FPR $=\mathrm{FP}/(\mathrm{FP}+\mathrm{TN})$) as $\tau$ varies, measuring how well the model separates co-visible from non-co-visible pairs independent of a fixed operating point; the corresponding ROC-AUC summarizes this trade-off.
The Precision-Recall (PR) curve instead plots Precision ($P =\mathrm{TP}/(\mathrm{TP}+\mathrm{FP})$) versus Recall (same as TPR) across thresholds and is often more informative under heavy class imbalance, as it focuses on the quality of predicted co-visible edges.
Finally, the $F_1$ score is the harmonic mean of Precision and Recall, $F_1=2\,\frac{\mathrm{Precision}\cdot\mathrm{Recall}}{\mathrm{Precision}+\mathrm{Recall}}$, and provides a single-number summary at a chosen threshold reflecting the balance between missing true co-visible pairs (FN) and introducing spurious co-visible edges (FP), which directly impacts downstream matching and graph construction.
For qualitative examples of IoU over various threhsolds see Fig.~\ref{fig:viz_graph_failures}.

\subsection{Strength Degree Prediction}
\label{supp:strenght_pred}
In addition to binary co-visibility classification, we optionally augment Co-VGGT with a lightweight strength regression branch that predicts a continuous overlap score in $[0,1]$. For each image pair, the dataloader provides a scalar target derived from the scene’s continuous relationship matrix when available (otherwise defaulting to the binary label). 
The regression branch mirrors the classification pathway: it is fed the same pairwise features computed from VGGT embeddings (under the chosen aggregation) and produces a scalar strength logit. 
In the multi-layer setting, this prediction is computed per selected layer and then aggregated across layers using the same strategy as classification—either a layer-wise MoE (softmax gating) or uniform averaging—so regression is consistent with the model’s layer reasoning mechanism rather than being attached only at the final stage. 
During training, the predicted strength is obtained via a sigmoid and optimized with a standard regression loss (MSE), combined with the classification objective using a tunable weight; when disabled, the model reduces to the original classification-only formulation.

As shown in Tab.~\ref{tab:regression_results}, the results are largely consistent with those obtained using the non-regression formulation. We attribute this to the intrinsic difficulty of the regression task. Estimating a precise score in the range [0,1] that reflects the overall percentage of overlap between two images is inherently ambiguous; for instance, values such as 0.2 or 0.3 could both be considered reasonable estimates, despite representing substantially different targets in a regression setting. Consequently, the regression formulation provides limited additional benefit.

Moreover, a large fraction of image pairs are non-covisible, which automatically assigns a regression target of 0. 
Such samples are largely non-informative during training and contribute little useful signal for the backpropagation process.

\begin{table}[t]
\centering
\caption{\textbf{Pair strenght regression.} Comparison between multiview and pairwise evaluation settings. Results are reported on Gibson for both settings.}
\label{tab:regression_results}
\scriptsize
\setlength{\tabcolsep}{4.5pt}
\renewcommand{\arraystretch}{1.05}

\begin{tabular}{lccccccc}
\toprule
\textbf{Method} 
& \textbf{MoE} 
& \textbf{Trainable} 
& \multicolumn{2}{c}{\textbf{Multiview}} 
& \multicolumn{2}{c}{\textbf{Pairwise}}  \\
\cmidrule(lr){4-5} \cmidrule(lr){6-7} 

& \textbf{Head} 
& \textbf{Parameters}
& \textbf{IoU* ($\uparrow$)} 
& \textbf{AUC ($\uparrow$)}
& \textbf{IoU* ($\uparrow$)} 
& \textbf{AUC ($\uparrow$)}\\

\midrule

\rowcolor{gray!10} Co-VGGT\textbf{ (Ours) }
& $\checkmark$ & 7.5M 
& \textbf{0.74} & \textbf{0.73} 
& \textbf{0.85} & 0.78 \\

Co-VGGT + \textit{regression}
& $\checkmark$ & 12M 
& \textbf{0.74} & 0.72
& 0.84 & \textbf{0.81} \\

\bottomrule
\end{tabular}

\end{table}

\begin{table}[t]
\centering
\scriptsize
\setlength{\tabcolsep}{2.5pt}
\renewcommand{\arraystretch}{1.15}
\caption{\textbf{Co-VGGT performance across difficulty levels (AUC).}
Left: edge-level difficulty defined by pairwise image overlap (Easy $\ge 50\%$, Med. $10$--$50\%$, Hard $<10\%$).
Right: graph-level difficulty defined by scene sparsity via average overlap (Easy $\ge 10\%$, Med. $4$--$10\%$, Hard $<4\%$).
Metrics report Graph AUC.}
\label{tab:difficulty_levels_auc}
\vspace{-0.2cm}

\begin{subtable}[t]{0.48\linewidth}
\centering
\caption{\textbf{Image overlap} (edge-level).}
\begin{tabular}{lcccc}
\toprule
\textbf{Method} & \textbf{Easy} & \textbf{Med.} & \textbf{Hard} & \textbf{Avg.} \\
\midrule
Co-VGGT (Pair) & 0.74 & 0.86 & 0.68 & 0.73 \\
Co-VGGT (Multi)  & \textbf{0.90} & \textbf{0.89} & \textbf{0.77} & \textbf{0.79} \\
\bottomrule
\end{tabular}
\end{subtable}
\hfill
\begin{subtable}[t]{0.48\linewidth}
\centering
\caption{\textbf{Scene sparsity} (graph-level).}
\begin{tabular}{lcccc}
\toprule
\textbf{Method} & \textbf{Easy} & \textbf{Med.} & \textbf{Hard} & \textbf{Avg.} \\
\midrule
Co-VGGT (Multi) & 0.94 & 0.88 & 0.68 & 0.71 \\
Co-VGGT (Pair)  & \textbf{0.96} & \textbf{0.90} & \textbf{0.77} & \textbf{0.79} \\
\bottomrule
\end{tabular}
\end{subtable}

\vspace{-0.2cm}
\end{table}

%\bibliographystyle{splncs04}
%\bibliographyS{main}

\end{document}